%% file: acl.tex
\pdfoutput=1
\documentclass[11pt]{article}
\usepackage[final]{acl}
\usepackage{times}
\usepackage{latexsym}
\usepackage[T1]{fontenc}
\usepackage{lipsum}   
\usepackage{tablefootnote}
\usepackage{graphicx}
\usepackage{amsmath,amssymb}
\usepackage{makecell}
\usepackage{tabularx, array}
\usepackage{graphicx}
\usepackage{multirow}
\usepackage{algorithm}
\usepackage{algorithmic}
\usepackage{booktabs}
\usepackage{fancyvrb}
\usepackage{tipa}
\usepackage[htt]{hyphenat}
\usepackage[textsize=tiny]{todonotes}

\usepackage{microtype}

\title{Exploring Sound Change Over Time: \\A Review of Computational and Human Perception}

\author{Siqi He \\
  Shanghai Jiao Tong University \\
  University of Heidelberg \\
  \texttt{hesiqiid@sjtu.edu.cn} \\\And
  Wei Zhao \\
  University of Aberdeen \\
  \texttt{wei.zhao@abdn.ac.uk} \\}

\date{}

\begin{document}
\maketitle
\begin{abstract}
Computational and human perception are often considered separate approaches for studying sound changes over time; few works have touched on the intersection of both. To fill this research gap, we provide a pioneering review contrasting computational with human perception from the perspectives of methods and tasks. Overall, computational approaches rely on computer-driven models to perceive historical sound changes on etymological datasets, while human approaches use listener-driven models to perceive ongoing sound changes on recording corpora. Despite their differences, both approaches complement each other on phonetic and acoustic levels, showing the potential to achieve a more comprehensive perception of sound change. Moreover, we call for a comparative study on the datasets used by both approaches to investigate the influence of historical sound changes on ongoing changes. Lastly, we discuss the applications of sound change in computational linguistics, and point out that perceiving sound change alone is insufficient, as many processes of language change are complex, with entangled changes at syntactic, semantic, and phonetic levels.

\end{abstract}

\input{introduction.tex}
\input{methods.tex}
\input{discussion.tex}
\input{conclusion.tex}

\section*{Acknowledgements}
We thank the anonymous reviewers for their thoughtful feedback that greatly improved the texts. 

\bibliography{acl}
\end{document}

%% file: introduction.tex
\section{Background}

There has been ongoing scholarly interest in sound change over time for decades.  
A popular historical sound change is 
the Great Vowel Shift \citep{lass1992phonology}, where the long vowel [i:] in 
Middle English shifted to a diphthong /\textipa{aI}/ 
in Modern English for example. It took place over time 
from the 15th to 18th centuries, and greatly changed the English vowel system. Other examples include the loss of voiceless velars like [\textipa{ç}] in Modern English \citep{dobson1968english}, reduction of consonant clusters like [kn] $\rightarrow$ [n] \citep{turville2020book}, and vowel reduction in unstressed syllables 
\citep{minkova2013historical}. Many ongoing sound changes took place in the 20th century. For instance, in American regional dialects, a notable shift such as [\textipa{2}] $\rightarrow$ [\textipa{E}] in the vowel system occurred in the Northern Cities around the mid-20th century \citep{wolfram_american_2016}.

While many works proposed computational approaches to perceive historical sound changes \citep{mielke2008emergence, dekker2018reconstructing, boldsen-paggio-2022-letters} and others suggested using the listener-driven model to perceive ongoing sound changes \cite{Janson1983SoundCI, Sanker2018ASO, Quam2021ImpactsOA}, few works explored the intersection of both computational and human perception.
The benefits of doing so can be substantial. Firstly, computational approaches perceive historical sound change by analyzing IPA transcriptions in etymological datasets, but these datasets lack acoustic features that human listeners/speakers can produce and perceive. Secondly, human perception observes ongoing changes through participation surveys over recording corpora, which lacks the considerations of acoustic and phonetic alignments between speakers that computational approaches can produce and perceive. Lastly, connections between etymological datasets and recording corpora are little explored; by combining both, one could conduct a comparative analysis of historical and ongoing sound changes, e.g., examining how historical changes impact ongoing changes. Thus, there is a need for a comparative review of computational and human perception.

In this work, we aim to fill the gap between two distant perception of sound change over time, with computational models on one hand and human observation on the other. To achieve this, we first review the tasks and methods from each perspective, and then present a unified view that combines both perception to explore sound change. Moreover, we discuss the connections of sound change to semantic and syntactic change, as well as the applications of sound change in computational linguistics.

%% file: methods.tex
\section{Computational Perception}
\label{sec:computing}
\paragraph{Sound change detection.} 
\citet{boldsen-paggio-2022-letters} connected semantic change detection with sound change detection and argued that diachronic distributional embeddings used for semantic change detection can track historical sound change. For lexical semantic change, diachronic word embeddings are guided by the distributional hypothesis suggesting that words that occur in similar contexts appear to have similar meanings. Interestingly, this idea also applies to phonology. Previous works showed that phonemes occurring in similar phonetic contexts likely belong to the same phonological class, demonstrating the applicability of the distributional hypothesis to phoneme embeddings \cite{mielke2008emergence, silfverberg2018sound}.

\citet{boldsen-paggio-2022-letters} proposed using phoneme embeddings trained on a historical Danish corpus to track sound changes over time.
Their approach compared the embeddings of a phoneme pair across different periods to observe sound changes. For instance, [p] $\rightarrow$ [b] is observed when the distance between the phoneme embeddings of [p] and [b] becomes smaller over time. The results showed a decrease in distance between three phoneme embedding pairs: [p] and [b], [t] and [d], [k] and [g] over time, meaning that their approach recognized the phonetic changes from voiceless plosives to their voiced counterparts in Danish. 

\paragraph{Phonetic alignment between cognate words.}
Phonetized cognate words consist of paired IPA transcriptions in two languages: either a proto-language and its descendant language or two descendant languages. Each transcription represents a sequence of phonemes. Translating a sequence of phonemes from one language to another can be framed as a machine translation task, as both execute a cross-lingual sequence-to-sequence task \citep{dekker2018reconstructing, fourrier2020comparing}.

For instance, \citet{fourrier2020comparing} proposed using both statistical and neural machine translation models to perform phoneme-level translations between cognate words. The models investigated include Moses \citep{koehn2007moses} and MEDeA \citep{luong-etal-2015-effective}. The languages considered include Latin, Italian, and Spanish. For evaluation, the generated translations were compared to the ground-truths through BLEU \cite{papineni-etal-2002-bleu}---which calculates the overlap of $n$-grams phonemes between translations and the ground-truths. The results showed that the phonetic translations between cognate words from the proto-language to a descendant language (or from a descendant language to another) are much better than those from a descendant language to the proto-language. Moreover, the results demonstrated the superiority of the statistical model over the neural MT model on small datasets, whereas the neural model showed a greater ability to handle many-to-one mappings from various proto-forms to the same descendant form. It is important to note that the ground-truth translations were collected by automatically phonetizing cognate word pairs via 
Espeak \citep{duddington20072015}, and the automatic phonetization process is prone to errors, meaning that comparing generated translations with the ground truths may lead to inaccurate model assessment.

\paragraph{Markedness of phonemes.} Markedness is a linguistic label separating common from less common phonemes in a phonological system. In English, the voiceless consonants [p], [t], and [k] are unmarked as they are more common compared to their voiced, marked counterparts [b], [d], and [g]. For vowels, peripheral high vowels such as [i] and [u] are marked while mid-central vowels like [\textschwa] are unmarked \citep{jakobson1968child, haspelmath2006against}.

\citet{ceolin2019modeling} proposed a probabilistic approach to model sound change by estimating the frequency of phonemes over time. Interestingly, they found that their approach could also recognize the markedness of a phoneme. Their approach was to estimate the frequency of each phoneme at a later time based on the frequencies of other phonemes observed at an earlier time through the split-merger process. The results showed that the unmarked phonemes at a later time appear to have higher frequencies compared to marked counterparts as postulated, meaning that their approach could separate unmarked from marked phonemes. We note that their approach considered neither phonetic nor acoustic features and was only evaluated on three phonemes in an artificial setup.

\paragraph{Sound convergence.} 
Unlike historical sound change, which takes place gradually over centuries, sound convergence is a process of ongoing sound change through which speakers adjust their speech to align acoustically and phonetically with other speakers \cite{natale1975social}. Research showed that native speakers often adapt their sound to non-native speakers in interactive environments \citep{giles1973accent, pardo2006phonetic, babel2010dialect,yu2013phonetic}. Recently, works by \citet{lewandowski2018vocal, wagner2021phonetic} showed that sound convergence can also occur in non-interactive settings. For instance, \citet{wagner2021phonetic} recruited 76 native Dutch speakers and a non-native speaker to read aloud a careful selection of words, and their speech was recorded. Importantly, although the native and non-native speakers have no interaction, the speech of non-native speakers was made available to the native speakers. This allows the native speakers to potentially adjust their speech. All the speech was transformed into acoustic features by using Praat \citep{boersma2011praat}. Acoustic features (e.g., vowel and fricative duration) of the native speakers are compared to the non-native speaker by calculating the difference-in-distance score over these acoustic features. The positive score indicates convergence, otherwise divergence. The results showed that (a) the Dutch native speakers show sound convergence to the non-native speaker in the scenario where interaction is minimal and (b) the degree of sound convergence is affected by how much native speakers think the speech by the non-native speaker is native-like. 

Despite being useful, relying on acoustic features to observe sound convergence has at least two limitations: Firstly, the convergence on the phonetic level is overlooked; for example, speakers who adjust their speech in terms of place and manner of articulation are not recognized as sound adaptation.
Secondly, the outcome of sound convergence is affected by the quality of acoustic features—which relies on the recording quality and the efficacy of computer tools to extract these features from speech.

\section{Human Perception}
\label{sec:human}
\paragraph{Perceptual similarity.} 
The work by \citet{goldinger1998echoes} introduced the perceptual similarity task that is concerned with how phonetic changes are received, processed and interpreted by listeners \citep{martin1981perception, sanker2018survey}. This approach is known as the listener-driven model of sound change, where a group of listeners heard recordings of a speaker's initial speech and the later speech (after the speaker listened to a target speech by another speaker). The listeners were then asked to determine which of the two recordings sounded closer to the target speech the speaker was exposed to. 

\paragraph{Sound convergence.}
\citet{wagner2021phonetic} employed the idea of  perceptual similarity to study sound convergence from native speech to non-native speech in Dutch. They recruited 16 listeners native in Dutch and asked them to perform the perceptual similarity task where the listeners heard participants' initial and later speech and had to choose which production sounded more similar to that of the model speaker non-native in Dutch. Moreover, the listeners were asked to rate the model speaker's speech in terms of how accented, comprehensible, and familiar it sounded. These ratings were included in the analyses to determine how they affected the degree of observed convergence. The results showed that the overall sound convergence score was slightly above the random chance, indicating a weakly perceived convergence in participants' speech after the target speech was exposed to them. Secondly, they found that several speech samples showed more sound convergence than others. Moreover, they noted that perceived convergence was affected by how strongly the model speaker's foreign accent was perceived. 

We note that although human perception can observe ongoing sound changes, listeners may misperceive the acoustic and phonetic features of a speaker, resulting in incorrect judgments of sound changes \citep{babel2010accessing, ohala2017phonetics, sanker2018survey}.

\section{A Unified Perspective}
\label{sec:intersections}

\paragraph{Computer-aided human perception.} 
Using computational methods can partly automate and possibly refine the human perception process of sound changes. This is because doing so allows for observing subtle changes on phonetic and acoustic levels, such as vowel duration shift and nasal place assimilation that are sometimes not obvious to perceive by listeners.
Additionally, combining computational and listener-driven methods would create a feedback loop where computational results could refine the human perception process and insights from  listeners could be used to improve computational models.

\paragraph{Cross-studying etymological datasets and recording corpora.}
Etymological datasets contain phonetic transcriptions that reflect historical sound changes, which are commonly used for computational models to observe changes over centuries. In contrast, recording corpora are a database for listeners to perceive ongoing sound changes. Despite their different aims, it is intriguing to know the influence of historical sound changes on ongoing changes. A potential idea is to start with collecting shared words in etymological datasets and recording corpora, and then inspect their phonetic similarity and difference. Note that unlike recording corpora, phonetic transcriptions do not include acoustic features; therefore, a comparative study of historical and ongoing sound changes at the acoustic level is not possible.  

%% file: discussion.tex
\section{Discussions}
\subsection{Connections to Other Changes}
\label{sec:connection}
While there have been many works on computational modeling of semantic and syntactic change \citep{hamilton-etal-2016-diachronic, schlechtweg-etal-2020-semeval, ma2024presence,ma-etal-2024-graph,merrill-etal-2019-detecting,krielke-etal-2022-tracing,chen2024syntactic}, they often lack connections to sound change. Such connections are crucial because many changes simultaneously affect multiple linguistic levels. A notable case is homograph where two words share the same spelling but have different meanings and pronunciations. Examples include ``present'' ([\textipa{\textprimstress prEz\textschwa nt}] vs. [\textipa{prI\textprimstress zEnt}]) and ``bow'' ([\textipa{baU}] vs. [\textipa{boU}]). Another case is grammaticalization---a process that incurs semantic, syntactic and phonetic changes. For example, ``going to'' grammaticalizes into ``gonna'', shifting from a verb to a future marker. This process changes the original meaning, impacts syntactic structure, and incurs phonetic reduction. To identify homographs and grammaticalization, it might be necessary to develop computational and human approaches that could model/observe changes across multiple linguistic levels at once.

\subsection{Applications in Computational Linguistics}

\paragraph{Phylogenetic Inference.} This task aims to reconstruct the evolutionary relationships among languages based on their shared linguistic features. For example, Proto-Indo-European, as the ancestral language, gives rise to many descendant languages within the Indo-European language groups such as Indo-Iranian, Germanic, and Celtic. Linguists construct a phylogenetic language tree by taking the ancestor language as the root and connecting it to descendant languages, based on the laws of sound changes over time \citep{hoenigswald1965language}. For instance, there exists a phoneme correspondence between High German [ts], Dutch [t], English [t], Swedish [t], and Icelandic [t], all of which are inherited from the proto-phoneme [*t] in their ancestry Proto-Germanic language group (where [*t] $\rightarrow$ [t] in High German). This phoneme correspondence is one of many reasons that these languages are the descendants of Proto-Germanic.

However, computer-based language phylogenies for major language groups like Dravidian \citep{kolipakam2018bayesian}, Sino-Tibetan \citep{sagart2019dated}, and Indo-European \citep{heggarty2023language} often rely on cognate sets from semantically aligned word lists across languages. \citet{campbell2008language} questioned the use of cognate sets for phylogenetic inference, as meanings in cognate words might undergo changes over time, resulting in the instability of a phylogenetic tree. Other works proposed reconstructing phylogenetic language trees using sound correspondences between cognate words instead of lexical cognates \citep{chacon2016improved, cathcart-2019-gaussian, chang-etal-2023-automating-sound, hauser-etal-2024-sounds}. For instance, \citet{hauser-etal-2024-sounds} presented a framework that first identifies phonetic alignment between cognate words using LingPy
and then uses BMrBayes \citep{ronquist2003mrbayes} and RAxML-NG \citep{kozlov2019raxml} to reconstruct phylogenetic trees. For evaluation, the generated phylogenetic trees are compared to the ground-truth Glottolog tree \citep{hammarstrom2019glottolog} by computing their topological distance via generalized quartet distance \citep{pompei2011accuracy}. The results showed that sound-based phylogenetic trees underperform cognate-based counterparts, i.e., that cognate-based trees are topologically closer to the gold Glottolog tree. This might be attributed to the lack of consideration for borrowing. For instance, two languages might not be related, although the phoneme sequences of their cognate words could be similar. Loanword is the example, where phonemes are borrowed from a third, unrelated language, rather than inherited from the proto-phoneme. 

\paragraph{Quality assessment of etymological datasets.}
Etymological datasets are a crucial resource for phylogenetic inference, low-resource machine translation, and historical linguistic tasks. Many such datasets have been made available and are automatically generated from various data sources. For instance, EtymWordNet \citep{de2014etymological} and CogNet \citep{batsuren2019cognet} are derived from WordNet across hundreds of languages, while EtymDB 1.0 \citep{sagot2017extracting} and 2.0 \citep{fourrier2020methodological} are sourced from Wiktionary across over two thousand languages. However, the quality of these datasets remains unclear. Firstly, many datasets use a loose definition of cognacy to enlarge data coverage. Secondly, the automatic processes used to generate these datasets are prone to errors. Therefore, there is a need to estimate the quality of these etymological datasets.

\citet{wettig2012using} proposed using the degree of phonetic alignment between cognate words as a measure of the internal consistency of an etymological dataset. They postulated that the more phonetically similar cognates words are, the better quality a dataset would be. For instance, the English word `house' and the German word  `Haus' are phonetically equivalent [\textipa{haUz}], implying that this cognate word pair is likely correct.
To achieve this idea, they use the Minimum Description Length, a dynamic programming algorithm, to calculate the cost of an optimal phoneme-level alignment between cognate word pairs for the Uralic language group. The alignment operates on phonetic features such as plosive/fricative and labial/dental. The challenge arises from the fact that phonemes inherited from the proto-phoneme may undergo sound changes over time, resulting in phonemes in one language potentially different from another. For evaluation, the generated alignments were not compared to the ground-truths due to the lack of gold phoneme-level alignments. Instead, their approach was evaluated in three scenarios: compression rates, rules of correspondence and imputation. 

Note that their phoneme-level alignments were not compared against the ground-truths. Thus, the efficacy of such a measure for estimating the quality of etymological datasets remains unclear. Moreover, their approach only considers one-to-one phoneme-level alignment and ignores one-to-many. As a result, cognate word pairs with one-to-many alignments (e.g., [\textipa{k\ae t}]  in `cat' and [\textipa{kats\textschwa}] in the German word `Katze') go unnoticed by their approach.

%% file: conclusion.tex
\section{Conclusions}
As two rarely connected disciplines, computational and human perception have their own interests, tasks and methods. However, we showed that these two aspects of perception benefit each other from the perspectives of methods and datasets. Additionally, we showed that the applications of sound change are manifold in computational linguistics, including phylogenetic inference and quality assessment of datasets. Despite these positive aspects, we argue that a unified perception of multi-faceted change is crucial, as many changes are entangled across phonetics, syntax and semantics.